\documentclass[sigconf]{acmart}
\usepackage{booktabs}       
\usepackage{graphicx}
\usepackage{xcolor}
\usepackage{multicol}
\usepackage{amsmath}
\usepackage{amsfonts}
\usepackage{multirow}
\usepackage{tabularx}
\usepackage{wrapfig}

\usepackage{tikz}
\usetikzlibrary{mindmap}
\usepackage{enumitem}

\usepackage{svg}
\usepackage{hyperref}
\usepackage{colortbl} 
\usepackage{
pgfplots,
tikz,
}
\usepgfplotslibrary{groupplots}

\pgfplotsset{
   compat=1.16,
   legend entry/.initial=,
}

\definecolor{bblue}{HTML}{4F81BD}
\definecolor{rred}{HTML}{FFB303}
\definecolor{ggreen}{HTML}{9BBB59}

\definecolor{igreen}{HTML}{579c35}
\definecolor{ppurple}{HTML}{9F4C7C}

\AtBeginDocument{%
  }

\acmConference[CIKM '25]{CIKM}{November 10-14,
  2025}{Seoul, Korea}
\acmISBN{978-1-4503-XXXX-X/2018/06}

\begin{document}

\title{Evaluating Differentially Private Generation\\
of Domain-Specific Text}
\author{Yidan Sun}
\orcid{0000-0002-3607-4963}
\affiliation{
  \institution{Imperial College London, Imperial Global Singapore}
  \country{Singapore}
}

\author{Viktor Schlegel}
\affiliation{
  \institution{Imperial College London, Imperial Global Singapore}
  \country{Singapore}
}

\author{Srinivasan Nandakumar}
\affiliation{
  \institution{Imperial College London, Imperial Global Singapore}
  \country{Singapore}
}

\author{Iqra Zahid}
\affiliation{
  \institution{Imperial College London, Imperial Global Singapore}
  \country{Singapore}
}

\author{Yuping Wu}
\affiliation{
  \institution{University of Manchester}
  \country{United Kingdom}
}

\author{Warren Del-Pinto}
\affiliation{
  \institution{University of Manchester}
  \country{United Kingdom}
}

\author{Goran Nenadic}
\affiliation{
  \institution{University of Manchester}
  \country{United Kingdom}
}

\author{Siew-Kei Lam}
\affiliation{
  \institution{Nanyang Technological University}
  \country{Singapore}
}

\author{Jie Zhang}
\affiliation{
  \institution{A*STAR}
  \country{Singapore}
}

\author{Anil A Bharath}
\affiliation{
  \institution{Imperial College London, Imperial Global Singapore}
  \country{Singapore}
}

\renewcommand{\shortauthors}{Sun, Schlegel, Nandakumar et al.}

\begin{abstract}

Generative AI offers transformative potential for high-stakes domains such as healthcare and finance, yet privacy and regulatory barriers hinder the use of real-world data. To address this, differentially private synthetic data generation has emerged as a promising alternative. In this work, we introduce a unified benchmark to systematically evaluate the utility and fidelity of text datasets generated under formal Differential Privacy (DP) guarantees. Our benchmark addresses key challenges in domain-specific benchmarking, including choice of representative data and realistic privacy budgets, accounting for pre-training and a variety of evaluation metrics. We assess state-of-the-art privacy-preserving generation methods across five domain-specific datasets, revealing significant utility and fidelity degradation compared to real data, especially under strict privacy constraints. These findings underscore the limitations of current approaches, outline the need for advanced privacy-preserving data sharing methods and set a precedent regarding their evaluation in realistic scenarios.\footnote{
Corresponding Authors: y.sun1@imperial.ac.uk and v.schlegel@imperial.ac.uk.}
\end{abstract}




\maketitle
\section{Introduction}

\begin{figure}[th]
\includegraphics[clip, trim=0.2cm 0.3cm 0.5cm 0cm, width=.9\columnwidth]{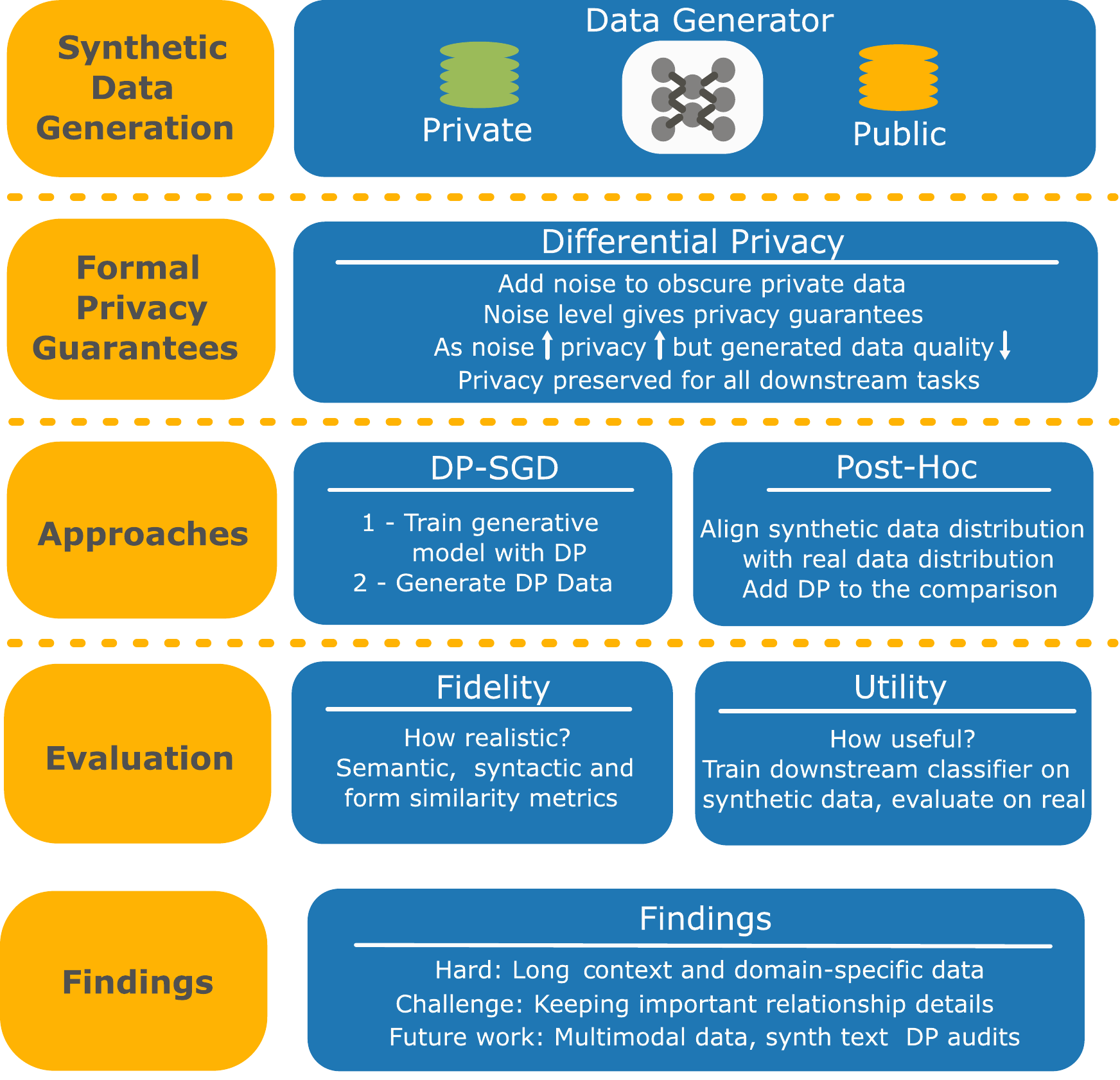}
\vspace{-1.0em}
\caption{We benchmark differentially private synthetic text generators using a variety of metrics and find that they struggle to produce realistic data from \textit{specialist domains}.}
\label{img:overview}
\vspace{-1.5em}
\end{figure}

The rapid advancement of Natural Language Processing (NLP) methods has seen success in an increasing number of tasks, including specialist domains, e.g., generating medical documents~\cite{Li2023Team:PULSARModels,Nagar2024UMedSum:Summarization,Binici2025MEDSAGE:Dialogues}, solving math problems~\cite{Cobbe2021TrainingProblems-cikm} or penetration-testing~\cite{deng2024pentestgpt}. However, their success hinges on the wide availability of training and benchmark data, which is exacerbated for potentially sensitive domain-specific data, due to regulatory privacy-related reasons. Without access to such domain-specific data, NLP systems' performance often deteriorates~\cite{Nagar2025LLMsExtraction} or becomes too unpredictable for domain experts to use~\cite{DellAcqua2023NavigatingQuality}.

Generative AI models offer a promising solution, providing a view on domain-specific data and its complexities by generating realistic synthetic datasets representing otherwise inaccessible sensitive data. Combining them with Differential Privacy (DP) \cite{Dwork20026}, the de-facto gold standard measure for formally quantifying the maximum disclosure risk associated with a data release, allows data holders to generate high-quality representative synthetic data to share with external AI researchers, while maintaining formal privacy guarantees~\cite{Schlegel2025GeneratingAhead}. However, the effectiveness of such methods has been primarily validated on either toy problems~\cite{Mattern2022DifferentiallySharing,Ochs2025PrivateModels} or open-domain datasets~\cite{Yue2023SyntheticRecipe,Xie2024DifferentiallyText}. This introduces two problems with respect to the truthful estimation of their efficacy:

Firstly---what we call \textbf{prior exposure}---generating synthetic text from public-domain datasets is comparatively simple, as publicly available data is likely to be found in the pre-training corpora of foundation models. 
This simplifies generation, as models tend to memorise their training data~\cite{Carlini2021ExtractingModels} thus leading to performance overestimates on various benchmarks~\cite{Ni2025TrainingNeed}, including synthetic text generation. Importantly, privacy leakage would be underestimated, as the generative model has access to the data not only during the synthesis process, where disclosure risk is controlled, but also during (un-controlled) pre-training~\cite{Tramer2024Position:Pretraining}.
Secondly---we coin this problem \textbf{representativeness}---the utilisation of general domain datasets, such as product reviews, as opposed to domain-specific datasets~\cite{Johnson2023MIMIC-IVDataset} excludes challenges associated with domain-specific data, such as domain-specific jargon~\cite{hudson1978jargon} or organisation- or country-specific work practices reflected in data. For instance, a real-world benchmark for clinical coding will incorporate hospital- and country-specific coding practices~\cite{Nguyen2023ACoding}; exposing them through data sharing, however, is challenging due to privacy concerns.

Generating domain-specific text data requires rigorous benchmarking and evaluation, an area that is currently underdeveloped, as existing methods do not share common benchmarks or evaluation protocols. To address this research gap, we propose a first effort to benchmark domain-specific textual dataset generation under formal DP guarantees. Specifically, we \emph{(a)} design our benchmark to include gated-access, domain-specific datasets to address the issues outlined above, \emph{(b)} introduce a rigorous and reproducible evaluation protocol aimed at precisely quantifying the performance of DP text generators; \emph{(c)} evaluate state-of-the-art approaches and showcase their struggles in realistic privacy-preserving data sharing scenarios. Lastly, we \emph{(d)} suggest future work and possible research avenues.

\vspace{-0.8em}
\section{Background \& Related Work}
\sloppy
Differential privacy is a formal method that provides an upper bound on the amount of information that can be inferred about a private dataset from a derived data release. Formally, a randomised mechanism $\mathcal{M}$ is $(\epsilon , \delta)$-differentially private, if for any of its two datasets $x, x'$, that differ in a single row, for each $S \subset Range(\mathcal M)$ the following inequality holds~\cite{Dwork20026}: \mbox{$Pr[\mathcal{M}(x) \in S] \leq exp(\epsilon) Pr[\mathcal{M}(x') \in S] + \delta$}.
 In other words: The difference in the probability densities for any possible subset of the output space for outputs of $M$ for $x$ and $x'$ differs by at most $exp(e)$ (in $1-\delta$\% of the cases, no guarantee for the rest)
thus giving an upper bound on the performance of a best-possible Membership Inference Attack (MIA). In the context of data synthesis, $x$ is a private dataset, $\mathcal M$ is a data generator and $\mathcal{M}(x)$ is a generated synthetic dataset. We are concerned with the setting of a ``trusted curator'' (e.g., a hospital) releasing a synthetic version of a private dataset, and looking to limit the ability of an adversary to infer whether any individual---represented by a single row in the private dataset---was part of the private dataset, by looking at the synthetic data (only).

Direct data manipulation (``rewriting'') is infeasible due to the high-dimensional, discrete nature of textual data, and because such anonymization efforts generally do not prevent re-identification~\cite{sweeney2002k}. Thus, existing text generation methods fall into two paradigms. First, \emph{DP training} approaches~\cite{Yue2023SyntheticRecipe,Mattern2022DifferentiallySharing,meeus2025canary} fine-tune generative models on ``control codes'' (e.g. class labels) using variants of DP-SGD~\cite{Abadi2016DeepPrivacy}, which privatizes training by clipping gradients and adding noise to bound each example’s influence. By post-processing, prompting these models with control codes yields DP synthetic data. Second, \emph{DP inference} methods privatize generation itself~\cite{koga2024privacy,Flemings2024DifferentiallyModels}, e.g., PATE~\cite{Papernot2018ScalablePATE}, which aggregates teacher predictions from disjoint private subsets. Because private data may be accessed at each generation step, privacy loss accrues per token, limiting scalability for large corpora~\cite{amin2024private}. 
A notable exception is AUG-PE~\cite{Xie2024DifferentiallyText}, which evolves random datasets toward the private distribution while privatizing the fitness function at relatively low privacy cost. 
In this paper, we evaluate one representative state-of-the-art method from each direction.

The evaluation of synthetic data generation varies significantly across modalities. For tabular data, comprehensive frameworks assess fidelity through structural similarity, distribution preservation, and global structure metrics \cite{pdpc2023,tabsyndex,bellinger2016beyond}. Text evaluation is fundamentally more challenging due to its high-dimensional, context-dependent nature. Most approaches measure distributions in embedding space using metrics like KL-divergence or MAUVE \cite{Pillutla2021MAUVE:Frontiers}, yet prior work has predominantly focused on ``easy-to-measure characteristics'' such as text length distributions \cite{Xie2024DifferentiallyText,Yue2023SyntheticRecipe}. Additionally, existing benchmarks typically use general-purpose datasets like sentiment analysis corpora or PubMed abstracts \cite{Blitzer2007BiographiesClassification,canese2013pubmed}. Our work addresses these limitations by introducing comprehensive evaluation metrics that go beyond surface-level properties to assess domain-specific utility and entity-level fidelity, while evaluating on specialist domains that provide more realistic assessments of privacy-preserving text generation.

\vspace{-0.5em}
\section{Benchmark Design}

We design the benchmark to address key challenges of evaluating differentially private synthetic text generators: 

\noindent\textbf{Addressing the challenges:} 
For \textbf{prior exposure}, we include gated access datasets (requiring e.g., Data Usage Agreements) to reduce the likelihood of the data appearing in LLMs' pre-training corpora used for data synthesis. 
Further, we utilise a mostly open language model~\cite{grattafiori2024llama} as a backbone of the evaluated methods, verifying that benchmark ``private'' data has not been exposed to the models during pre-training\footnote{They might have still encountered the data during closed-source post-training.}. For \textbf{representativeness} We use challenging domain-specific datasets from the (bio)-medical, clinical and legal domain.

\noindent\textbf{Choice of $\epsilon$:}
While there is no consensus how much $\epsilon$ is ``enough''~\cite{Lee2011HowPrivacy-cikm}, high values provide increasingly lower protections~\cite{Dwork2019DifferentialEpsilons}. In practice\footnote{e.g., as recommended by NIST: \url{https://www.nist.gov/blogs/cybersecurity-insights/differential-privacy-future-work-open-challenges}} $e\leq5$ or even $e\leq1$ is considered a strong privacy guarantee, which guides our choice: $\epsilon \in \{0.5,1,2,4\}$.

\noindent\textbf{Evaluation Protocol:} Our benchmark evaluates both utility and fidelity at different $\epsilon$-levels. \emph{Utility} quantifies how useful the synthetic data is for the real downstream application task. We do this by training \emph{multiple} downstream classification models on the synthetic data and evaluating their performance on a (held-out) test set of original data, assessing how well synthetic data supports downstream tasks. Meanwhile, \emph{fidelity} is the assessment of how similar synthetic data is with respect to the original dataset. Our implementation supports the quality evaluation of synthetic text, across various metrics. 
For surface-level similarity, we report BLEU~\cite{Papineni2001} and, METEOR~\cite{Banerjee2005METEOR:Judgments}, which capture n-gram and sequence overlap between real and synthetic sentences. BERTScore~\cite{Zhang2019BERTScore:BERT} and Universal Sentence Encoder (USE)~\cite{Cer2018UniversalEnglish} cosine similarity are used to evaluate semantic alignment. 
For these reference-based metrics, we use a ``many reference'' evaluation approach, where each synthetic sentence is compared against the entire pool of original sentences. 
Corpus-level fidelity evaluation includes measuring differences in distributions between original and synthetic data: MAUVE~\cite{Pillutla2021MAUVE:Frontiers},  recognised Named Entities and text lengths evaluate semantic, content and structural similarity, respectively. To quantify the effect of varying privacy noise levels, we compute these metrics for each synthetic dataset at varying~$\epsilon$.

\noindent\textbf{Methods \& Datasets:}  
We compare two state-of-the-art differentially private text generators: DP-Gen (DP-SGD generation from~\citet{Yue2023SyntheticRecipe}) and AUG-PE~\cite{Xie2024DifferentiallyText}, representing distribution alignment. 
We evaluate three healthcare datasets---\textsc{HoC} \cite{DBLP:journals/bioinformatics/BakerSGAHSK16} (cancer hallmark identification in scientific literature), \textsc{N2C2'08} \cite{uzuner2009recognizing} (obesity and co-morbidity recognition in clinical discharge summaries), and \textsc{PsyTAR} \cite{Zolnoori2019} (adverse drug effect detection in social media posts). We further include a financial dataset, \textsc{DMSAFN} \cite{daniel2023financialnews}, and \textsc{AsyLax} \cite{barale-etal-2023-automated} to extend coverage to legal reasoning. 
As summarized in Table~\ref{tab:datasets}, the benchmark includes long documents (\textsc{HoC}, \textsc{N2C2'08}), gated-access corpora (\textsc{PsyTAR}, \textsc{N2C2'08}), small datasets challenging to train on (\textsc{N2C2'08}), and multi-label settings with many labels (\textsc{HoC}, \textsc{N2C2'08}, \textsc{PsyTAR}).

\begin{table}[htbp]
\vspace{-1.0em}
\centering
\caption{Benchmark characteristics: Number of labels ($|C|$), dataset size ($|D|$), average/90th percentile token length $|\overline L|$ and data access.}
\vspace{-0.8em}
\label{tab:datasets}
\begin{tabular}{lcccc}
\toprule
\textbf{Dataset} & \textbf{$|C|$} & \textbf{$|D|$} & \textbf{$|\overline{L}|$} & \textbf{Access mechanism} \\
\midrule
\textsc{HoC} & 10 & 10301 & 37/58 & free on HuggingFace\\
\textsc{N2C2'08} & 16 & 620 & 1985/3070 & DUA \& manual approval \\
\textsc{PsyTAR} & 7 & 5102 & 19/34 & DUA \\
\textsc{DMSAFN} & 3 & 3876 & 30/50 & free on HuggingFace \\
\textsc{AsyLax} & 3 & 7999 & 2326/3442& free on HuggingFace\\
\bottomrule
\end{tabular}
\vspace{-1.6em}
\end{table}
\section{Results \& Analysis}

\begin{table}[t]
\centering
\small
\caption{Random/majority guess baselines, and average/best downstream classification F1 scores for three models per dataset across different privacy budgets ($\epsilon$). 
}
\label{tab:text-utility-v2}
\vspace{-0.85em}
\resizebox{0.99\columnwidth}{!}{\begin{tabular}{lcccccc}
\toprule
\textbf{Method} &  & $\epsilon = \infty$ & $\epsilon = 4$ & $\epsilon = 2$ & $\epsilon = 1$ & $\epsilon = 0.5$ \\
\midrule
\rowcolor{black!10}
\textsc{HoC} & 
\multicolumn{6}{c}{
(\textbf{Random}: 3.7, \textbf{Majority}: 9.1, \textbf{Original}: 71.6/74.3)
} \\
DP-Gen (avg / best) &  & 52.0 / 58.6 & 15.5 / 17.9 & 14.4 / 15.7 & 11.0 / 11.5 & 13.6 / 14.9 \\
AUG-PE (avg / best) &  & 15.8 / 19.7 & 10.3 / 12.1 & 7.5 / 8.6 & 8.1 / 9.4 & 6.7 / 10.0 \\
\midrule
\rowcolor{black!10}
\textsc{N2C2'08} &
\multicolumn{6}{c}{
(\textbf{Random}: 46.1, \textbf{Majority}: 53.2, \textbf{Original}: 73.1/87.7)
} \\
DP-Gen (avg / best) &  & 54.9 / 58.2 & 53.2 / 53.2 & 47.8 / 53.2 & 53.7 / 54.7 & 57.1 / 61.3 \\
AUG-PE (avg / best) &  & 55.9 / 57.6 & 55.8 / 60.9 & 54.7 / 57.7 & 56.8 / 60.0 & 55.7 / 56.9 \\
\midrule
\rowcolor{black!10}
\textsc{PsyTAR} &
\multicolumn{6}{c}{
(\textbf{Random}: 25.4, \textbf{Majority}: 41.8, \textbf{Original}: 80.7/82.1)
} \\
DP-Gen (avg / best) &  & 69.5 / 70.1 & 41.6 / 44.0 & 40.1 / 42.5 & 40.8 / 42.3 & 39.2 / 39.1 \\
AUG-PE (avg / best) &  & 65.9 / 67.3 & 62.1 / 62.6 & 58.4 / 60.4 & 56.0 / 57.4 & 45.7 / 48.1 \\
\midrule
\rowcolor{black!10}
\textsc{DMSAFN} &
\multicolumn{6}{c}{
(\textbf{Random}: 30.5, \textbf{Majority}: 41.1, \textbf{Original}: 76.8/95.2)
} \\
DP-Gen (avg / best) &  & 62.7 / 91.6 & 47.5 / 70.1 & 46.5 / 65.8 & 49.4 / 69.6 & 48.0 / 65.0 \\
AUG-PE (avg / best) &  & 51.0 / 65.6 & 51.0 / 65.9 & 51.7 / 70.9 & 47.9 / 61.3 & 50.0 / 65.9 \\
\midrule
\rowcolor{black!10}
\textsc{AsyLax} &
\multicolumn{6}{c}{
(\textbf{Random}: 38.4, \textbf{Majority}: 51.4, \textbf{Original}: 64.9/69.0)
} \\
DP-Gen (avg / best) &  & 60.2 / 61.5 & 34.8 / 36.2 & 34.9 / 39.3 & 47.6 / 48.3 & 37.0 / 48.0 \\
AUG-PE (avg / best) &  & 50.5 / 51.6 & 49.3 / 52.1 & 51.5 / 51.5 & 51.9 / 52.9 & 51.4 / 51.5 \\
\bottomrule
\end{tabular}}
\vspace{-1.4em}
\end{table}

Table~\ref{tab:text-utility-v2} reveals substantial utility degradation in text generation. Importantly, even without privacy constraints ($\epsilon = \infty$), synthetic data fails to fully match real-data baselines suggesting  that the \textbf{evaluated methods cannot fully capture domain-specific complexity}. Under strong privacy constraints and averaged across all datasets ($\epsilon \leq 4$), the average of models' performance stays at around 50\% of real-data performance, strikingly independently of $\epsilon$. Discounting for random/majority baseline performance and calculating the overall improvement over baselines (capped by performance on real data, similar to relative gain by~\citet{mattern2022limits}), lowers this score to 55\%/21\% without privacy guarantees for DP-Gen/Aug-PE, and to an improvement of at most 28\%/52\% with $\epsilon \leq 4$. Only when looking at the best performing model on each dataset, the expected utility-privacy trade-off becomes visible (e.g., 28\%, 26\%, 23\%, 21\%, 15\% retention for AUG-PE at $\epsilon \in \{\inf, 4,2,1,0.5\}$, with DP-Gen dominating over AUG-PE without privacy constraints (62\% vs 28\% improvement over the baseline). This finding validates our benchmark design, specifically the use of multiple classifier baselines, to obtain more precise data utility estimates. Looking at per-dataset performance, gated-access datasets present significant challenges---for example \textsc{N2C2'08} has the worst baseline-adjusted utility. This yields evidence towards our hypothesis that evaluating on non-publicly available data results in a more realistic performance estimate.

\begin{table}[ht]
\vspace{-0.3em}
\centering
\small
\caption{MAUVE ($\mathcal{M}$) as well as entity ($\mathcal{N}$) and text length ($\mathcal{L}$) distribution divergences (fidelity) for different approaches and privacy budgets ($\epsilon$) across datasets. 
}
\label{tab:text-fidelity-v2}
\vspace{-0.85em}
\resizebox{0.48\textwidth}{!}{\begin{tabular}{p{1.1cm}@{}c@{\hskip.1cm}c@{\hskip.1cm}c@{\hskip.1cm}c@{\hskip.1cm}c}
\toprule
\multirow{2}{*}{\textbf{Method}} 
& $\epsilon = \infty$ 
& $\epsilon = 4$ 
& $\epsilon = 2$ 
& $\epsilon = 1$ 
& $\epsilon = 0.5$ \\
& $\mathcal{M}\!\uparrow/\mathcal{N}\!\downarrow/\mathcal{L}\!\downarrow$ 
& $\mathcal{M}\!\uparrow/\mathcal{N}\!\downarrow/ \mathcal{L}\!\downarrow$ 
& $\mathcal{M}\!\uparrow/\mathcal{N}\!\downarrow/ \mathcal{L}\!\downarrow$ 
& $\mathcal{M}\!\uparrow/\mathcal{N}\!\downarrow/ \mathcal{L}\!\downarrow$ 
& $\mathcal{M}\!\uparrow/\mathcal{N}\!\downarrow/ \mathcal{L}\!\downarrow$ \\
\midrule
\rowcolor{black!10} \textsc{HoC} &
\multicolumn{5}{c}{ (\textbf{Original}: 0.99/1.04/0.004)} \\
DP-Gen & 0.65/2.10/0.06 & 0.18/4.33/0.40 & 0.16/4.35/0.417 & 0.14/4.37/0.44 & 0.12/4.55/0.47 \\
AUG-PE & 0.01/2.87/1.49 & 0.01/3.19/1.40 & 0.01/3.29/1.35 & 0.01/4.04/1.25 & 0.01/4.85/1.25 \\
\midrule
\rowcolor{black!10} \textsc{N2C2'08} &
\multicolumn{5}{c}{ (\textbf{Original}: 0.99/0.70/0.17)} \\
DP-Gen & 0.42/1.57/0.75 & 0.02/9.98/1.37 & 0.02/9.91/1.40 & 0.02/9.58/1.42 & 0.02/9.86/1.54 \\
AUG-PE & 0.02/8.55/3.72 & 0.03/8.50/3.80 & 0.02/8.31/3.87 & 0.02/8.53/3.63 & 0.02/8.78/3.42 \\
\midrule
\rowcolor{black!10} \textsc{PsyTAR} &
\multicolumn{5}{c}{ (\textbf{Original}: 0.99/0.86/0.007)} \\
DP-Gen & 0.61/2.00/0.03 & 0.42/2.71/0.03 & 0.34/2.92/0.05 & 0.37/3.55/0.03 & 0.35/4.01/0.03 \\
AUG-PE & 0.02/4.31/3.21 & 0.02/5.02/3.27 & 0.02/5.06/3.23 & 0.02/5.35/3.34 & 0.02/6.29/3.38 \\
\midrule
\rowcolor{black!10} \textsc{DMSAFN} &
\multicolumn{5}{c}{ (\textbf{Original}: 0.86/0.71/0.012)} \\
DP-Gen & 0.21/1.72/0.06 & 0.18/1.94/0.02 & 0.16/2.03/0.04 & 0.15/2.22/0.03 & 0.12/2.36/0.06 \\
AUG-PE & 0.01/3.25/2.36 & 0.01/3.87/2.35 & 0.01/4.19/2.34 & 0.01/5.77/2.24 & 0.01/3.98/2.30 \\
\midrule
\rowcolor{black!10} \textsc{AsyLax} &
\multicolumn{5}{c}{ (\textbf{Original}: 0.98/0.11/0.004)} \\
DP-Gen & 0.03/1.42/0.43 & 0.01/3.82/6.41 & 0.01/4.03/5.95 & 0.01/8.06/3.58 & 0.01/7.66/1.23 \\
AUG-PE & 0.01/6.15/1.41 & 0.01/5.76/1.41 & 0.01/6.42/1.92 & 0.01/7.43/6.12 & 0.01/7.36/1.41 \\
\bottomrule
\end{tabular}
}
\vspace{-1.7em}
\end{table}
Table~\ref{tab:text-fidelity-v2} shows that fidelity also deteriorates, following similar trends of utility. MAUVE scores between real and synthetic data are close to zero across all domains, with entity overlap divergence particularly pronounced in \textsc{HoC} and \textsc{N2C2'08}, possibly due to domain-specificity and closed-access nature of the datasets. Interestingly, \textsc{AsyLex} again mirrors this pattern: low MAUVE and high NER divergence indicate failure to preserve legal argumentation structure, even when surface vocabulary is retained. Here, similarly, \textsc{N2C2'08} and \textsc{AsyLax} prove to be the most challenging datasets, which we attribute to the combination of long context, domain specificity, and, for \textsc{N2C2'08}, gated access.

\begin{figure}[t]
\includegraphics[clip, trim=0.2cm 0.3cm 0.5cm 0cm, width=0.95\columnwidth]{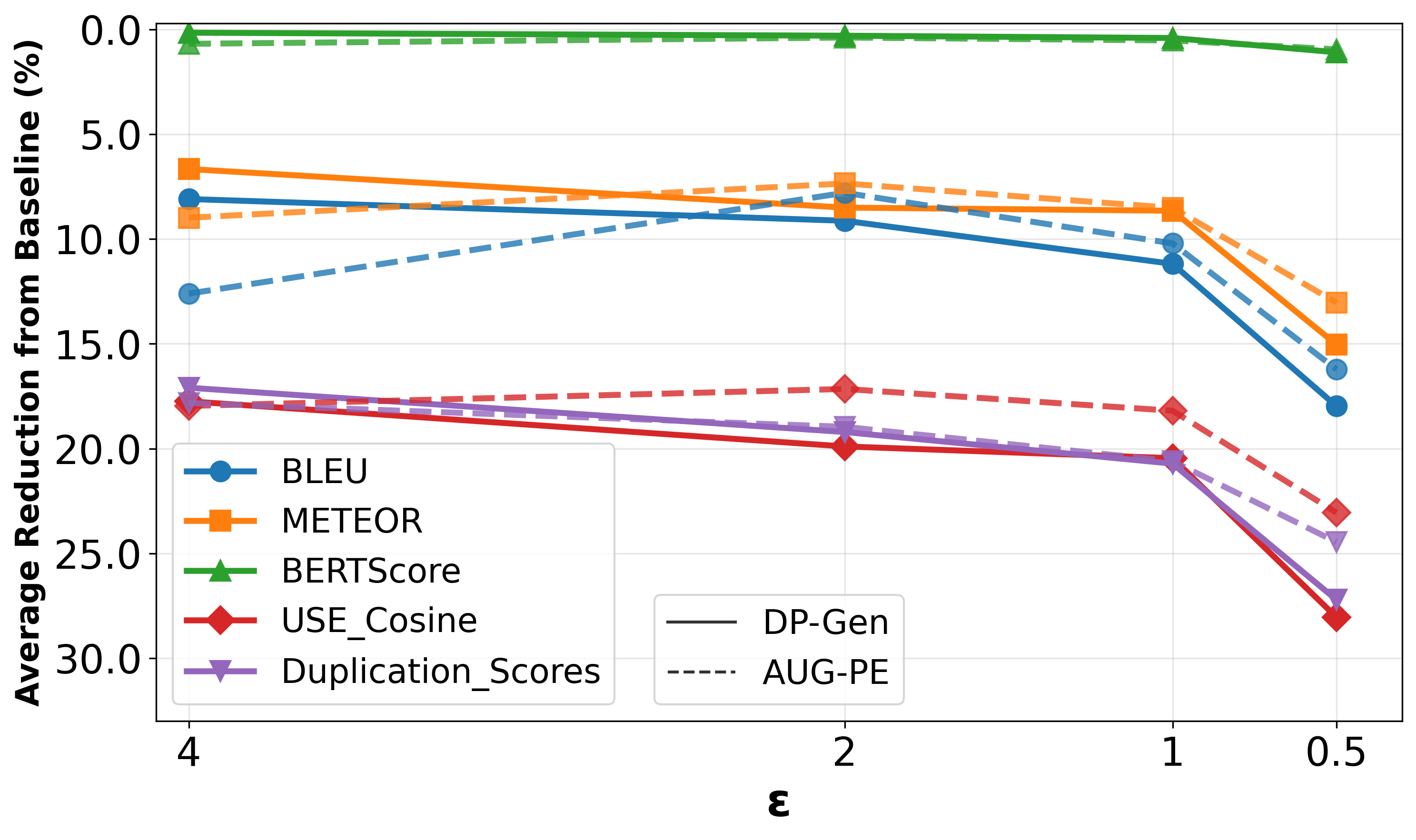}
\vspace{-1em}
\caption{Results for ``many reference'' metrics of DP-Gen and AUG-PE generated data at different $\epsilon$ levels, expressed as relative decrease compared to non-privately synthesised data. As noise increases, the quality deteriorates.}
\label{img:lexical-scores}
\end{figure}

Our comparison of lexical and semantic measures of AUG-PE and DP-SGD finetuned generators, as seen in Figure~\ref{img:lexical-scores}, 
shows that data quality (compared to non-privately synthesised data) progressively decreases, with similar declines for both methods. This privacy-fidelity trade-off is in line with the literature, with stronger privacy leading to lower data quality~\cite{Yue2023SyntheticRecipe,Xie2024DifferentiallyText,Mattern2022DifferentiallySharing}.

Overall, text generated by fine-tuned models exhibits good utility and fidelity (as evidenced by high utility at $\epsilon = \infty$, high MAUVE scores and low text length divergences), but the addition of DP noise deteriorates the quality of the generated data significantly, as noted by~\citet{Xie2024DifferentiallyText}, especially on ``unfamiliar'' domain-specific data\footnote
{which demands larger parameter updates that are clipped, noised; long contexts force smaller batch size, further raising the noise-to-signal ratio}. Conversely, data generated by AUG-PE is of initially low quality (low utility/MAUVE scores at $\epsilon=\infty$), but the elegant noising of histogram counts introduces much less noise in the process, thus deteriorating the quality less thereafter. This phenomenon is visualised in Figure~\ref{fig:text-example-v2}: while DP-Gen text maintains the overall structure (e.g. text length, general sense of having adverse drug effects), the AUG-PE example misses the social-media tone. With enough good candidates, the example would have been filtered out by the private evolution mechanism, suggesting that the embedding model underlying the histogram mechanism cannot faithfully capture the nuances of domain-specific data. 
Notably, we observe markedly lower performance than reported previously~\cite{Yue2023SyntheticRecipe,Xie2024DifferentiallyText,Mattern2022DifferentiallySharing}, \textbf{suggesting that evaluating DP text generators on open-domain, simple datasets overestimates their performance for real use-cases}.

\begin{figure}[!t]
\footnotesize
\setlength{\tabcolsep}{3pt}
\renewcommand{\arraystretch}{1.05}
\begin{tabularx}{0.47\textwidth}{|ccX|}
\hline
\textbf{Method} & $\epsilon$ & \textbf{Text} \\
\hline
\textbf{Original} & $-$ & \emph{Almost no libido at all, Utterly useless.} \\
\rowcolor{black!10}
\textbf{AUG-PE} & $0.5$ & \emph{You might experience blurry vision, weird food smells, a dry mouth, frequent trips to the bathroom, dry throat, occasional dizziness, feelings of restlessness and aches, and unsteadiness.} \\
\textbf{DP-Gen} & $0.5$ & \emph{I have had a few side effects.} \\
\hline
\end{tabularx}
\vspace{-1.1em}
\caption{Example generations for the label ``Adverse Drug Effects'' from the \textsc{PsyTAR} dataset.}
\vspace{-1.1em}
\label{fig:text-example-v2}
\end{figure}

\section{Conclusion and Future Work}

We present a benchmark design for synthetic text generation under formal privacy guarantees, alongside preliminary empirical results on healthcare text data. The results reveal significant challenges in generating high-quality synthetic data: state-of-the-art models face substantial performance deterioration when they are applied to domain-specific datasets increasingly so under high privacy constraints. This highlights the limitations of approaches which focus on foundation-models which are pre-trained on general domain data. Our findings emphasise the need for domain-specific approaches that faithfully represent domain-specific data while preserving privacy. Our work is a first move towards creating a standardised benchmark that can catalyse the progress in privacy-preserving synthetic data generation for high-stake applications. To accelerate this, we release our code and report additional metrics and scores, omitted in this paper due to space constraints, in a public git repository: \url{https://github.com/ImperialGlobalSingapore/synth-data}.

Future work will focus on expanding this benchmark: we aim to support multimodal data generation to evaluate the preservation of complex relationships between different data types e.g., clinical text reports and medical images \cite{Johnson2019MIMIC-CXRReports} and more advanced corpus evaluation metrics encompassing discourse structure and human assessments. 
Finally, while in this work we have focussed primarily on the data quality, we will also develop stronger membership inference attacks that leverage changes in these metrics to provide a more realistic assessment of privacy risks. To better validate privacy guarantees, our future work will introduce diagnostic datasets specifically designed for auditing membership inference attacks (MIA) on synthetic data~\cite{Meeus2025TheText}. This targeted approach will allow researchers that propose novel DP text generation methods to audit implementation correctness without generating large volumes of synthetic data to train shadow classifiers~\cite{Guepin2023SyntheticData-cikm}.

\vspace{-0.5em}
\section*{Acknowledgement}
This research is part of the IN-CYPHER programme and is supported by the National Research Foundation, Prime Minister’s Office, Singapore under its Campus for Research Excellence and Technological Enterprise (CREATE) programme. We are grateful for the support provided by Research IT in the form of access to the Computational Shared Facility at The University of Manchester and the computational facilities at the Imperial College Research Computing Service (DOI: \url{https://doi.org/10.14469/hpc/2232}). We also thank the anonymous CIKM reviewers for their feedback that helped us improve the paper further.

\section*{GenAI Usage Disclosure}
Some authors have used generative AI tools (GenAI) to polish grammar and suggest wordings. One author has relied on GenAI to propose an (empty) table format, which was later manually populated by corresponding results. No part of this paper has been entirely generated by GenAI.

\bibliographystyle{ACM-Reference-Format}
\bibliography{references-viktor,sample-base}


\end{document}